\documentclass[10pt,twocolumn,letterpaper]{article}

\usepackage[pagenumbers]{cvpr} 

\usepackage{graphicx}
\graphicspath{{figures/}}  
\usepackage{amsmath}
\usepackage{amssymb}
\usepackage{booktabs}
\usepackage{amsmath}
\usepackage{amsfonts}
\usepackage{array}
\usepackage{multirow}
\usepackage{longtable}
\usepackage{float}
\usepackage{listings}
\usepackage{cutwin} 
\usepackage{subcaption} 
\usepackage{lipsum} 
\usepackage{placeins} 
\usepackage{siunitx} 
\usepackage{booktabs} 
\usepackage[pagebackref,breaklinks,colorlinks]{hyperref}

\usepackage[capitalize]{cleveref}
\crefname{section}{Sec.}{Secs.}
\Crefname{section}{Section}{Sections}
\Crefname{table}{Table}{Tables}
\crefname{table}{Tab.}{Tabs.}

\def\cvprPaperID{8969} 
\def\confName{CVPR}
\def\confYear{2026}

\begin{document}

\title{Diffusion-Driven Inter-Outer Surface Separation for Point Clouds with Open Boundaries}

\author{Zhengyan Qin\\
Hong Kong University of Science and Technology (HKUST)\\
Clear Water Bay, Kowloon, Hong Kong.\\
{\tt\small 
zlqin@connect.ust.hk}
\and
Liyuan Qiu\\
Hong Kong Applied Science and Technology Research Institute (ASTRI)
\\
Photonics Centre, 2 Science Park E Ave, Hong Kong.\\
{\tt\small liyuanqiu@astri.org}
}
\maketitle
\begin{center}
    \vspace{-0.75em}
    {\small \textbf{Code:} \url{https://github.com/lambo131/diffusion-scatter-algorithm}}
\end{center}

\begin{abstract}
We propose a diffusion-based algorithm for separating the inter and outer layer surfaces from double-layered point clouds, particularly those exhibiting the "double surface artifact" caused by truncation in Truncated Signed Distance Function (TSDF) fusion during indoor or medical 3D reconstruction. This artifact arises from asymmetric truncation thresholds, leading to erroneous inter and outer shells in the fused volume, which our method addresses by extracting the true inter layer to mitigate challenges like overlapping surfaces and disordered normals. We focus on point clouds with \emph{open boundaries} (i.e., sampled surfaces with topological openings/holes through which particles may escape), rather than point clouds with \emph{missing surface regions} where no samples exist. Our approach enables robust processing of both watertight and open-boundary models, achieving extraction of the inter layer from 20,000 inter and 20,000 outer points in approximately 10 seconds. This solution is particularly effective for applications requiring accurate surface representations, such as indoor scene modeling and medical imaging, where double-layered point clouds are prevalent, and it accommodates both closed (watertight) and open-boundary surface geometries. Our goal is \emph{post-hoc} inter/outer shell separation as a lightweight module after TSDF fusion; we do not aim to replace full variational or learning-based reconstruction pipelines.

\end{abstract}

\section{Introduction}
\label{sec:intro}

\begin{figure*}[t]
    \centering
    \subfloat{\includegraphics[width=0.7\linewidth]{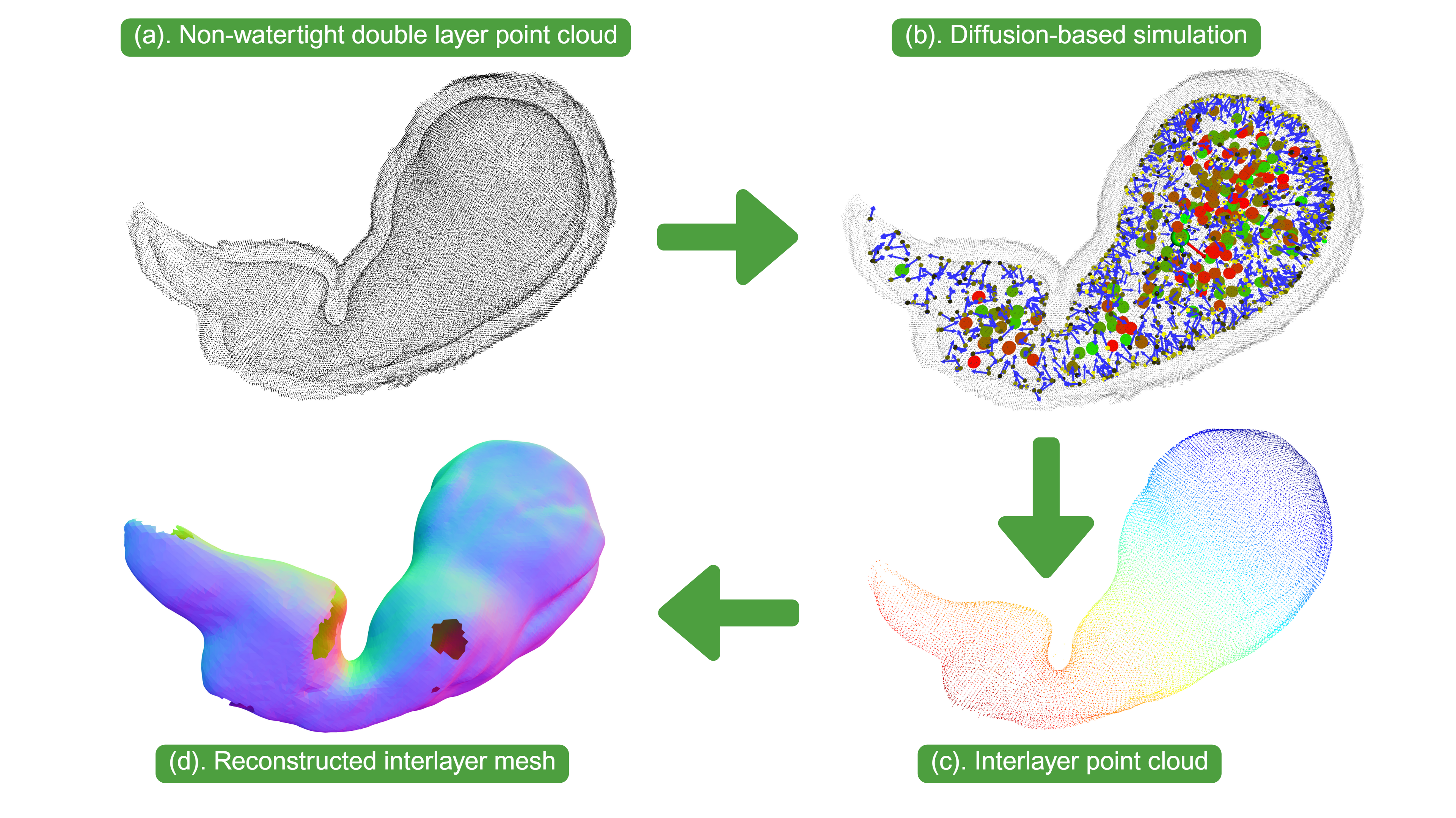 }}
    \caption{Diffusion-based inter-layer point cloud separation and reconstruction. (a) TSDF-fused stomach point cloud with a double-surface artifact (inter/outer shells) and topological openings. (b) Particle diffusion simulation: the initial spawn point (red sphere), newly generated spawn points (blue spheres), collided interior points (blue points), and the escape boundary enclosing the cloud (dashed sphere). (c) Extracted inter-layer point cloud. (d) Example surface mesh reconstructed from the extracted inter-layer points (see Sec.~\ref{sec:surface_reconstruction}).}
    \label{fig:pipeline}
\end{figure*}

The Truncated Signed Distance Function (TSDF) fusion algorithm~\cite{curless1996volumetric} (and related methods, such as KinectFusion~\cite{newcombe2011kinectfusion} and Open3D~\cite{zhou2018open3d}) can produce artifacts in 3D reconstructions, including fake double-layered structures. These artifacts emerge when the algorithm misinterprets noisy or incomplete data, yielding a mesh with two closely spaced surfaces---an inter and an outer layer---that are connected at edges or seams. For example, reconstructing thin objects like walls or sheets may lead to the creation of dual surfaces due to noise or misalignment, rather than a single cohesive one. Topologically, this manifests as a single connected 2D manifold, often with boundaries. While humans can intuitively differentiate the inter and outer layers using visual cues, traditional algorithms face significant challenges in separating them, exacerbated by high point cloud densities or ambiguous TSDF values near connecting edges. Such fake layer issues render surface reconstruction highly challenging, necessitating robust methods to distinguish the true (real) layer from the erroneous inter (phantom) layer. Advanced topological analysis or geometric constraints are thus essential for accurate layer separation.

The elimination of these phantom layers has broad applications across diverse fields. In robotics and SLAM~\cite{dougherty2025hybrid}, it enhances mapping accuracy for navigation in cluttered environments. In AR/VR and indoor 3D reconstruction~\cite{li20193d}, it ensures precise scene modeling for immersive experiences. Medical imaging~\cite{schmidt2024tracking} benefits from cleaner organ or tissue surfaces, improving diagnostic visualization. Similarly, industrial applications~\cite{he2020real}, such as quality inspection and digital twins, rely on artifact-free models for reliable metrology.

To distinguish inter and outer surfaces in these fake double-layered 3D models---without requiring watertight geometries---we propose a physics-inspired algorithm based on particle diffusion via random walks within the model. Our focus is \emph{post-hoc} separation of double-shell point clouds produced by TSDF fusion, especially when normals are unreliable or missing and the surface may contain openings. We do not aim to outperform full variational or learning-based reconstruction pipelines; instead, we provide a simple separation module that can be applied directly to point clouds.

\textbf{Terminology note.} In this paper, we use \emph{open-boundary} (sometimes loosely referred to as \emph{non-watertight}) to mean a sampled surface with \emph{topological openings} (holes) where diffusion particles can escape (as in Fig.~\\ref{fig:opensphere}), while the remaining surface is still supported by point samples. This is different from a point cloud with \emph{missing surface regions} (no samples in a region), for which any method---including ours---cannot reliably infer the absent interior surface and may misclassify nearby outer points.
\section{Related Work}

Surface reconstruction from point clouds or other sparse data representations has been a longstanding challenge in computer vision and graphics, with applications spanning 3D modeling, robotics, and medical imaging. Traditional approaches can be broadly categorized into geometric element generation methods, which focus on local or global fitting of primitives, and physical simulation-inspired algorithms that leverage volumetric or ray-based techniques. More recent advancements incorporate neural implicit representations for enhanced robustness to noise and incompleteness. In this section, we review key methods in these areas, highlighting their strengths and limitations, particularly in handling noise, thin structures, holes, and non-uniform sampling.

\subsection{Geometric Element Generation Methods}
Early geometric methods emphasize the creation of meshes through local connectivity or filtration processes. The \textbf{Poisson Reconstruction Method}~\cite{kazhdan2006poisson} and its screened variant~\cite{kazhdan2013screened} solve a Poisson equation to derive a watertight implicit surface from oriented point clouds. These approaches are robust to noise and produce smooth, closed surfaces, making them suitable for noisy scans; however, they tend to merge thin layers and bridge holes due to their emphasis on watertightness, which can distort fine details. \textbf{Screened Poisson Reconstruction} improves upon this by incorporating constraints to preserve sharp features, though it still requires careful parameter tuning and may bridge gaps unintentionally.

\textbf{Conformal alpha shape filtration}~\cite{cazals2005conformal,giesen2006conformal} and related Delaunay-based techniques filter tetrahedra using an $\alpha$-radius to extract manifold surfaces. These methods excel at capturing thin gaps with small $\alpha$ values and avoid over-smoothing, but they are sensitive to parameter selection and often bridge layers at holes, leading to topological artifacts. Similarly, the \textbf{Ball-Pivoting Algorithm} (BPA)~\cite{bernardini2002ball} "rolls" a sphere to connect points into triangles within a specified radius, preserving thin layers if the radius is smaller than inter-layer gaps and providing explicit control over such gaps. Nonetheless, BPA struggles with non-uniform point densities and bridges layers at holes, limiting its applicability to irregularly sampled data.

\textbf{Advancing Front Techniques}~\cite{farestam1995framework} incrementally expand a mesh front from seed points, adapting well to local densities and preserving sharp edges. While effective for complex geometries, they face challenges with topology and often bridge layers at holes. \textbf{Voronoi-based methods}, such as Cocone~\cite{dey2003tight}, leverage Voronoi diagrams for theoretically guaranteed reconstructions, preserving thin features in densely sampled regions with minimal artifacts. However, they are computationally intensive and prone to bridging at holes in sparse areas.

\textbf{Region Growing and Clustering approaches} segment points into local clusters before meshing each independently, which helps prevent global layer merging and avoids bridging across disparate regions. Despite these advantages, they may still bridge within holes and require accurate pre-segmentation, which can be error-prone. \textbf{Moving Least Squares (MLS) combined with meshing}~\cite{lim2007mls} smooths points via local polynomial fitting to denoise data and enhance mesh quality, but it exacerbates merging of thin layers and bridging at holes, particularly in noisy inputs.

\textbf{Marching Cubes}~\cite{lorensen1998marching} voxelizes a signed distance function (SDF) or isosurface and generates a polygonal mesh by marching through grid cells. This method handles complex topologies and fills holes smoothly, making it a cornerstone for volumetric data; however, it merges layers if the voxel resolution exceeds gap sizes and introduces artifacts at holes or sharp edges.

\subsection{Physical Algorithms}
Physical simulation-inspired methods treat reconstruction as a rendering or isosurface extraction problem. 
\textbf{Ray Casting}~\cite{roth1982ray,795213} traces rays through a volume to detect intersections, enabling detailed surface rendering, but it also applies to water-tight 3D models. 


\subsection{Neural Implicit Methods}
Recent neural approaches represent surfaces implicitly via deep networks, offering flexibility for incomplete or noisy data. Methods like Neural Radiance Fields (NeRF)~\cite{mildenhall2021nerf} and Deep Signed Distance Functions (DeepSDF)~\cite{park2019deepsdf} learn continuous surface representations from multi-view images or points. These techniques infer missing data effectively and handle noise well, but they may merge thin layers unless explicitly trained on datasets with similar gaps, and they often smooth over holes, leading to over-generalized surfaces. 
Besides, recent works have addressed the double surface artifact in TSDF fusion, such as RoutedFusion~\cite{weder2020routedfusion}, which employs a routing network for real-time depth map fusion to mitigate noise and erroneous shells, and FineRecon~\cite{stier2023finerecon}, which introduces a depth-aware feed-forward network with resolution-agnostic TSDF supervision for detailed reconstructions. However, these learning-based approaches typically demand substantial computational resources, including GPU-accelerated training and inference, and often require dataset-specific fine-tuning to generalize across diverse scenes, limiting their applicability in resource-constrained or real-time settings. Importantly, they operate by \emph{changing the fusion/reconstruction pipeline}, whereas our method is a \emph{post-hoc} shell separation step applied to an already reconstructed point cloud.

In summary, while traditional geometric and physical methods provide efficient and theoretically grounded solutions, they often struggle with holes, thin structures, and noise without extensive parameter tuning. Variational/PDE-based approaches (e.g., Laplace-operator formulations) provide a principled alternative but typically require solving a global optimization/PDE on a volumetric grid or a mesh. In contrast, our method is a lightweight, point-cloud-only random-walk procedure (no PDE solve, no mesh connectivity requirement) aimed at separating inter/outer shells in TSDF outputs.

\section{Gas Diffusion and Absorption Model}
\label{sec:physicalmodel}
The gas diffusion process is a fundamental physical phenomenon ensuring that gas particles achieve a uniform distribution within a confined volume over time, driven by random thermal motion. In the context of a three-dimensional domain with complex geometry, this process can be leveraged to analyze the structure of non-watertight 3D models, such as point clouds representing objects with holes or perforations.

Consider a vacuum domain enclosed by a non-watertight boundary, such as a point cloud derived from a 3D scan. When a small volume of gas particles is introduced, these particles undergo diffusion, exploring the accessible volume through random trajectories. Over time, particles collide with every point on the inter surface of the domain’s boundary (i.e., the inter layer of the enclosing wall), providing a statistical map of surface interactions. In non-watertight models, particles may escape through holes, potentially colliding with the outer surface of the boundary.

To model this behavior and distinguish inter from outer surfaces, we introduce an \textit{escape boundary} exterior to the domain. Particles that reach this boundary through holes are captured and removed, simulating diffusion and minimizing their interactions with the outer surface. This approach combines diffusion-driven exploration with diffusion-based escape, enabling robust identification of inter and outer surfaces in topologically complex, non-watertight 3D reconstructions. By tracking collision frequencies and escape events, our method infers geometric properties, such as holes or cavities, critical for applications in 3D computer vision and computational geometry.

\subsection{Theoretical Intuition (Qualitative)}
Our simulation can be viewed as a Markovian random walk inside the interior volume, with reflections at the sampled boundary.
Under standard diffusion intuition, long trajectories tend to explore the accessible volume in an approximately ergodic manner, and the empirical distribution of boundary ``hits'' concentrates on the interior boundary that is reachable from the spawn points.

We emphasize that we do not provide a formal guarantee, but the following assumptions help explain when the method works well: (i) the inter surface is sampled sufficiently densely relative to the step size, (ii) openings are bounded (so escaping particles are absorbed by the escape boundary before accumulating many outer-surface collisions), and (iii) spawn points are inside the interior volume.
Under these conditions, collision frequency is dominated by the inter surface, while outer-surface hits remain rare.
\section{Implementation}
\label{sec:implement}
\subsection{Overview}
\begin{figure}[h]
    \centering
    \includegraphics[width=1.0\linewidth]{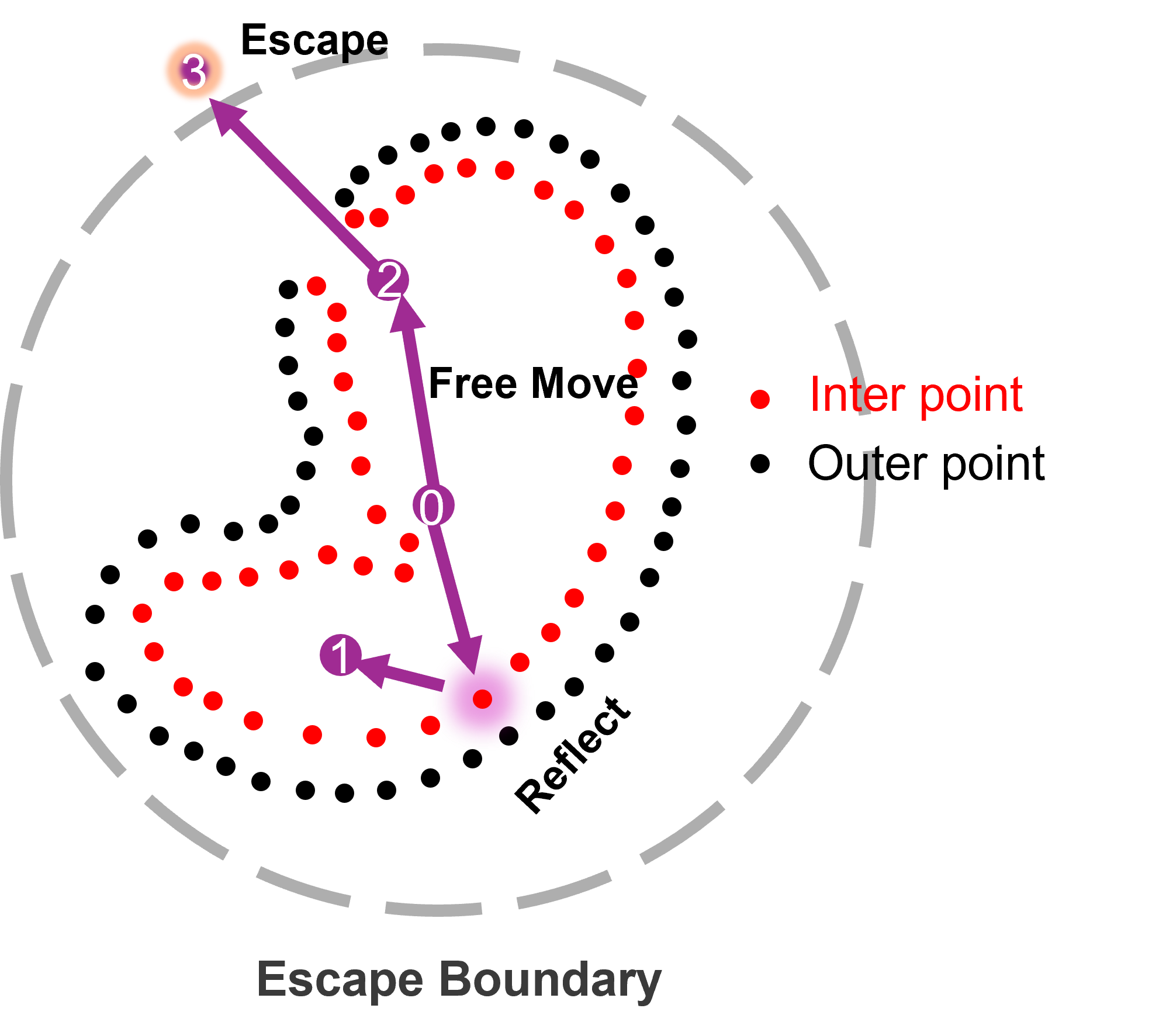}
    \caption{Diffusion algorithm visualization. The point cloud represents a hollow object with inter surface points (red) and outer surface points (black). Purple balls marked 0, 1, 2, and 3 represent the initial spawn point, reflected point, free-moving point, and escape point, respectively. The gray dashed sphere is the escape boundary sphere centered at the initial spawn point (0).}
    \label{fig:DiffusionAlgorithm}
\end{figure}

The diffusion algorithm simulates the movement of a ball within a hollow 3D point cloud to identify inter points of the object's geometry. The point cloud consists of two sets of points: inter surface points (red points in Figure~\ref{fig:DiffusionAlgorithm}) and outer surface points (black points in Figure~\ref{fig:DiffusionAlgorithm}). The algorithm initializes a simulation ball at a spawn point(the purple ball marked with 0 in Figure~\ref{fig:DiffusionAlgorithm}), tracks its collisions with the point cloud, and generates new spawn points to explore the geometry. The simulation terminates when the simulation ball collision number reaches a predefined limit or exits the escape boundary sphere which encloses the entire point cloud as shown by the gray dashed sphere in Figure~\ref{fig:DiffusionAlgorithm}).

\subsubsection{Initialization}
The simulation begins by setting up the environment and parameters as follows:
\begin{enumerate}
    \item \textbf{Point Cloud Loading}: Load the raw point cloud data of a hollow object as a NumPy array, representing 3D points that define the object's geometry. The point cloud consists of interior surface points (total number $ N_{\text{inter}} $), representing the interior surface, and outer surface points (total number $ N_{\text{outer}} $), representing the exterior surface. The total point count is $ N_t = N_{\text{inter}} + N_{\text{outer}} $. The unit length of point cloud $R_0$ as the average length of the nearest point cloud distance.
    \item \textbf{Spawn Point Initialization}: Designate an initial spawn point manually (purple sphere marked "0" in Figure~\ref{fig:DiffusionAlgorithm}) inside the point cloud, serving as the original spawn point of the simulation(usually the geometry center of the point cloud).
    \item \textbf{Escape Boundary Setup}: Define an escape boundary sphere centered at the bounding box center with a radius large enough to enclose the entire point cloud (gray dashed sphere in Figure~\ref{fig:DiffusionAlgorithm}).
    \item \textbf{Simulation Parameters Setup}:
    \begin{itemize}
        \item \( R_{\text{ball}} \): Radius of the simulation ball.
        \item \( L_{\text{max}} \): Maximum distance the simulation ball moves in a single step if no collision occurs.
        \item \( R_{eff} \): Effective collision radius, defined as \( R_{eff}  = R_{\text{ball}} + \text{collision margin} \).
    \end{itemize}
\end{enumerate}

\subsubsection{Simulation Process}
The simulation iteratively traces the trajectory of a simulation ball as it moves, collides within the point cloud , or escapes from the point cloud. The process is as follows:
\begin{enumerate}
    \item \textbf{Simulation Ball Initialization}:
    \begin{itemize}
        \item Spawn a simulation ball with radius \( R_{\text{ball}} \) at a spawn point, chosen with probability $p_0$ for a random spawn point and $ 1-p_0 $ for the initial spawn point. If no generated spawn points exist, default to the initial spawn point.
        \item Assign a random initial direction vector to the simulation ball.
    \end{itemize}
    \item \textbf{Simulation Ball Trajectory}:
    \begin{itemize}
        \item The ball moves in a straight line for a distance up to \( L_{\text{max}} \) unless a collision occurs.
        \item Upon collision with a point in the point cloud, compute a reflected direction based on the local surface geometry, adding a small random perturbation to simulate realistic scattering (e.g., transition from point 0 to 1 in Figure~\ref{fig:DiffusionAlgorithm}).
        \item Track the number of steps (discrete movements) and collisions. Terminate the ball if it reaches predefined limits for steps or collisions.
        \item If the ball exits the point cloud (e.g., through a hole), it may collide with outer surface points(which can cause the false point cloud collision) or reach the escape boundary sphere (e.g., transition from point 2 to 3 in Figure~\ref{fig:DiffusionAlgorithm}). Terminate the ball movement upon reaching the escape boundary sphere.
    \end{itemize}
    More details about the ball movement and collision is written in Section~\ref{sec:collision_detection}.
    \item \textbf{Dynamic Spawn Point Generation}: Generate new spawn points during the simulation based on collision process (see Section~\ref{sec:spawn_point_generation} for details). 
    \item \textbf{Iterative Simulation}: Upon termination of a simulation ball (due to step/collision limits or escape), initialize a new simulation ball and repeat the process from \textbf{Simulation Ball Initialization}, incorporating generated spawn points and logged data.
\end{enumerate}
\subsubsection{Simulation Termination}
\label{sec:simulation_termination}
The simulation terminates when either of two conditions is met: the total number of simulation balls reaches a predefined maximum, or the duplication rate $R_{dup}$, as defined in Section~\ref{sec:result}, reaches 0.99 for 10 consecutive iterations. These conditions ensure the algorithm stops when sufficient exploration is achieved or when further iterations yield minimal new information.

\subsubsection{Surface Reconstruction}
\label{sec:surface_reconstruction}
After the diffusion process, we obtain an extracted inter-layer point set. To visualize results as a mesh (e.g., Fig.~\ref{fig:pipeline}(d) and Fig.~\ref{fig:MethodContrast}), we reconstruct a surface using Poisson surface reconstruction on the extracted inter-layer points. Since open boundaries and locally sparse regions can lead to unreliable implicit values, small holes may appear in the reconstructed mesh; in our pipeline these holes primarily reflect either (i) true openings in the input or (ii) regions where inter-layer points are under-sampled after separation.
\subsection{Collision Detection}
\label{sec:collision_detection}
During each simulation step, the simulation ball can either move freely, collide with a point, or escape the boundary sphere:
\begin{itemize}
    \item \textbf{Free Movement}: Use Open3D’s \texttt{kdtree.search\_radius\_vector\_3d (current\_ball\_position, L\_max +$ R_{eff}$ )} 
    to identify cloud points within the ball’s potential movement volume ($ L_{\text{max}} + R_{eff}  $). If no points are found, move the ball by \( L_{\text{max}} \) in its current direction.
    \item \textbf{Collision}: If points are found within the movement range, identify real collision points by checking for overlap with the ball’s path (using effective collision radius $R_{eff}$). Select the closest point as the collision point. If no real collision occurs (e.g., no positive roots in the collision equation), move the ball by \( L_{\text{max}} \).
    \item \textbf{Escape}: After movement, check if the simulation ball’s distance from the center of the escape boundary sphere exceeds the boundary's radius. If so, mark the simulation ball as escaped and terminate this simulation iterative.
\end{itemize}

\subsection{Spawn Point Generation}
\label{sec:spawn_point_generation}
 
To improve exploration of the 3D point cloud's internal geometry, new spawn points are generated during the simulation. As the simulation ball collides with the point cloud at multiple points, the two most recent collision points are identified, and their midpoint is designated as a new spawn point. This method leverages the likelihood that the midpoint of two consecutive collisions lies within the internal volume of the 3D model, facilitating effective exploration of its geometry.

For non-convex objects, the midpoint heuristic can occasionally produce candidates outside the interior volume (false spawn points). To reduce this effect, we apply a lightweight validation: a candidate spawn point is rejected if (i) its nearest-neighbor distance to the point cloud is below a small threshold (indicating it lies on/too near the surface) or (ii) a short probe random walk starting from this point escapes the boundary within a small number of steps (indicating it is likely outside).

To maintain diversity and avoid redundancy, a maximum of 200 spawn points is enforced. Additionally, each new spawn point is checked for proximity to existing spawn points; if it is too close (based on a predefined distance threshold), it is discarded, and the algorithm waits for the next collision to generate a new spawn point.
\subsection{Output Information}
\label{sec:output_information}
The algorithm outputs a Python dictionary containing the following parameters: \{Collide($i$)\}, the set of point cloud points collided with by the simulation ball at the $i$-th simulation step; \{New($i$)\}, the set of unique (non-repeating) collided points at the $i$-th step; \{Dup($i$)\}, the set of duplicate collided points (points hit multiple times) at the $i$-th step; $C_{outer}$, the number of outer surface points collided with by the simulation ball; $C_{inter}$, the number of interior surface points collided with; and $N_{nescap}$, the number of times the simulation ball escapes the designated escape boundary.

\section{Results}
\label{sec:result}

The performance of the diffusion-based algorithm is evaluated using several metrics, particularly focusing on the duplication rate and the detection of inter and outer layer points in a 3D point cloud model. 

For the $i$-th simulation step, the duplication rate is defined as:
\[
R_{dup}(i) = \frac{|\{ \text{Dup}(i) \}|}{|\{ \text{Collide}(i) \}|},
\]
where $|\{ \text{Dup}(i) \}|$ is the number of duplicate collided points (points hit multiple times) and $|\{ \text{Collide}(i) \}|$ is the total number of collided points at step $i$. The diffusion process terminates when $R_{dup}(i) \geq 0.99$ for 10 consecutive simulation steps, indicating that most collisions involve previously hit points, suggesting near-complete exploration of the inter layer.

During the simulation, the total number of steps $i$ is the sum of steps where the simulation ball escapes the model ($N_{escape}$) and where it remains within the model ($N_{nescap}$), i.e., $i = N_{escape} + N_{nescap}$. For a watertight inter layer, $N_{escape} = 0$, as the simulation ball cannot exit. However, in a non-watertight model, $N_{escape} > 0$. We use the criterion $N_{escape} > 5$ to classify the point cloud model as non-watertight, accounting for potential simulation errors.  

The detected point cloud, $C_{detect}$, comprises the inter layer points, $C_{inter}$, and outer layer points, $C_{outer}$, such that $C_{detect} = C_{inter} + C_{outer}$. The detection rate for the inter layer is defined as:
\[
R_{inter} = \frac{C_{inter}}{N_{inter}},
\]
where $N_{inter}$ is the total number of inter layer points. Ideally, $R_{inter} = 1$ when all inter layer points are detected. Similarly, the detection rate for the outer layer is:
\[
R_{outer} = \frac{C_{outer}}{N_{outer}},
\]
where $N_{outer}$ is the total number of outer layer points. Ideally, $R_{outer} = 0$ when no outer layer points are detected. 
\subsection{Performance in Watertight 3D Model}
\label{sec:performance_watertight}
\begin{figure}[t]
    \centering
    \subfloat{\includegraphics[width=0.8\linewidth]{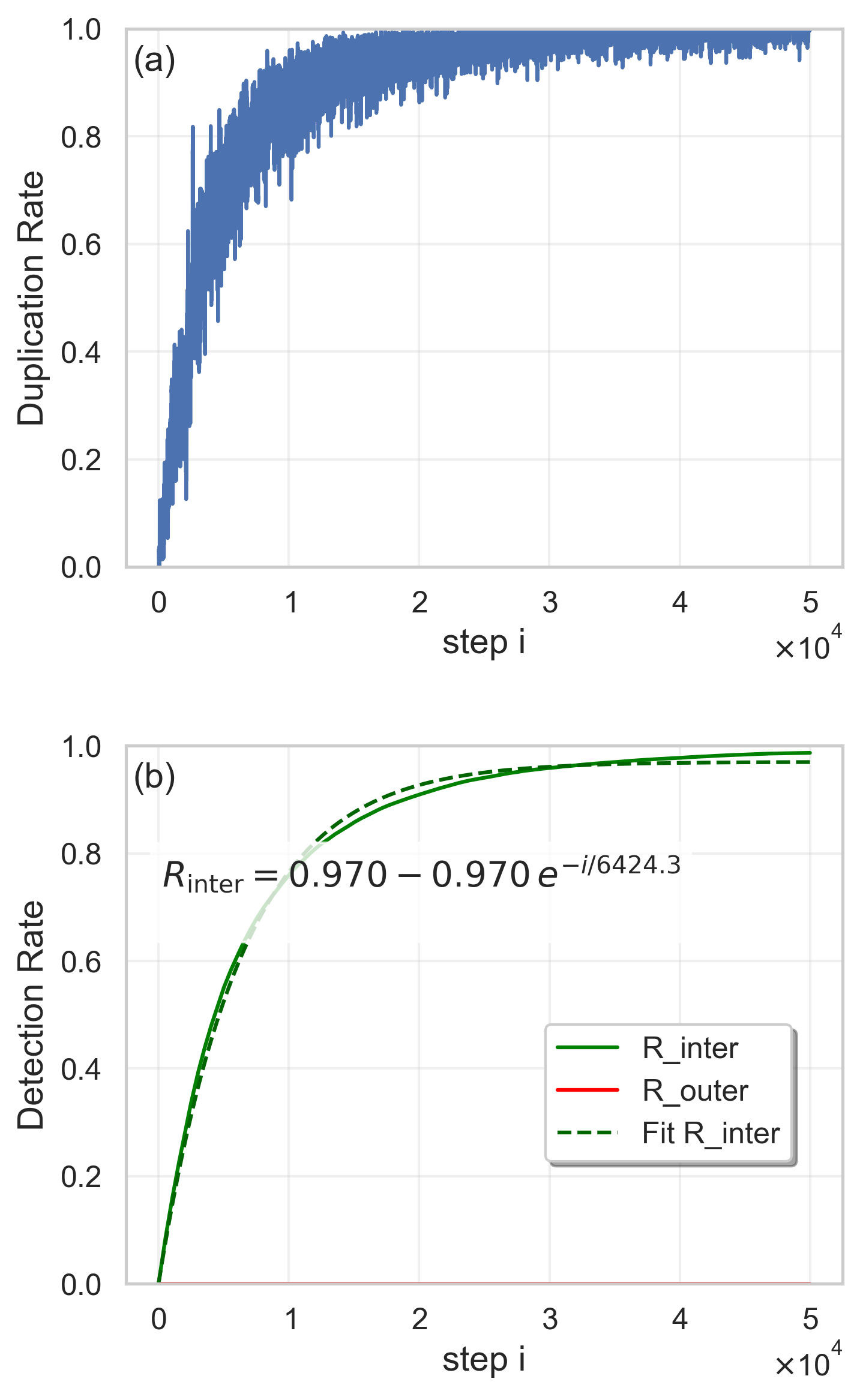}}
    \caption{ Performance metrics of the diffusion-based algorithm of closed double layer ball 3D model(20000 inter layer points and 20000 outer layer points) with $R_{ball}=2 R_0$.  
    (a) Duplication rate $R_{dup}(i)$ versus step $i$. 
    (b) The $R_{outer}$(red line) and $R_{inter}$(green line) versus step $i$. The $R_{inter}$ curve is fitted by the exponential saturation model $R_{inter}(i)=A_0\,(1-\exp(-i/\tau))$ with $A_0=0.970$ and $\tau=6424.3$.}
    \label{fig:performance}
\end{figure}

In this simulation of a watertight double layer ball 3D
mode, we set $R_{ball}=2R_0$, $collision \; margin=0.5 R_{ball}$, $R_{eff} = 1.5 R_{ball}$, and $L_{max}=50R_0$, $p=0.999$, and limit the number of steps to 50 and the collision limit to 5. The metrics are shown in Fig.~\ref{fig:performance} for $R_{dup}$ versus step $i$ and $R_{inter}, R_{outer}$ versus step $i$.

To quantify the performance of our diffusion-driven algorithm in identifying inter surfaces of non-watertight point clouds, we define $R_{\text{inter}}(i)$ as the proportion of correctly identified inter surface points after $i$ particle simulation iterations. The convergence of $R_{\text{inter}}$ is modeled using an exponential saturation function:
\begin{equation}
R_{\text{inter}}(i) = A_0 \left( 1 - \exp\left(-\frac{i}{\tau}\right) \right),
\label{eq:rinter}
\end{equation}
where $A_0 = 0.970$ represents the saturation level (the asymptotic coverage of inter points) and $\tau = 6424.3$ is the time constant governing the rate of convergence. This model reflects the cumulative effect of particle collisions, where the algorithm progressively covers the inter surface through diffusion, with performance stabilizing as more particles are simulated.

In practice, $A_0$ can be less than 1 due to finite simulation budget, locally sparse sampling, and geometric configurations where random walks have low probability of visiting certain regions (e.g., sharp concavities). We fit $(A_0,\tau)$ to the measured $R_{inter}(i)$ curve (least squares) and report them as a compact summary of coverage and convergence speed.

Results show that $R_{dup}(i)$ approaches 1 as fewer inter layer points remain undetected, indicating convergence of the diffusion process. 


\subsection{The simulation result in non-watertight 3D model}
\begin{figure}
    \centering
    \includegraphics[width=0.8\linewidth]{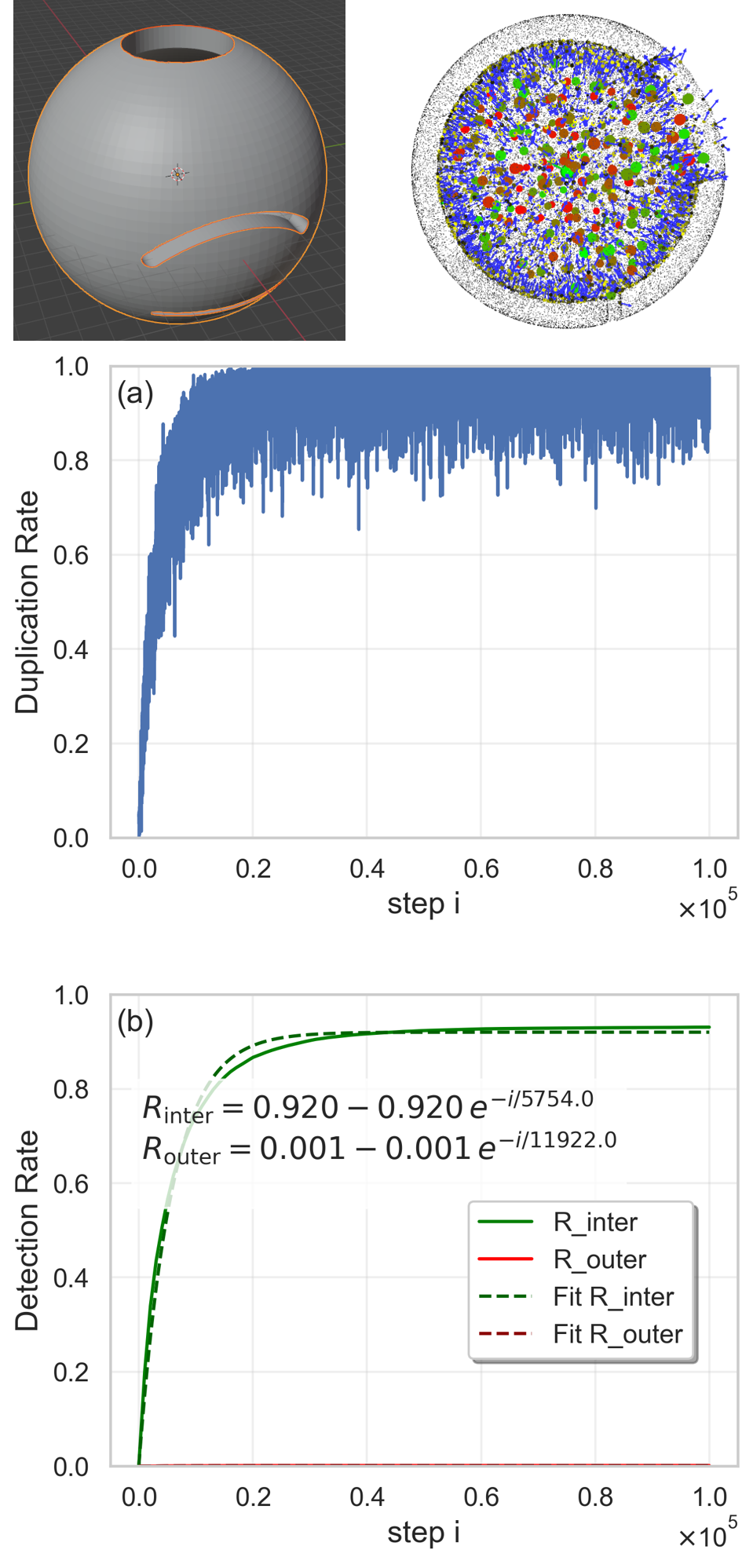}
    \caption{ Performance metrics of the diffusion-based algorithm of opened double layer ball 3D model(20000 inter layer points and 20000 outer layer points) with $R_{ball}=3 R_0$.  
    (a) $N_{escape}$ versus step $i$. 
    (b) The $R_{outer}$(red line) and $R_{inter}$(green line) versus step $i$. The $R_{inter}$ and $R_{outer}$ are fitted by exponentially saturation function.}
    \label{fig:opensphere}
\end{figure}

The simulation in an open-boundary 3D ball model is summarized in Fig.~\ref{fig:opensphere}, showing $N_{escape}$ versus step $i$ and $R_{inter}, R_{outer}$ versus step $i$. Additionally, only a small proportion (0.001 in Fig.~\ref{fig:opensphere}(b)) of outer-layer points are detected, confirming the algorithm’s effectiveness in prioritizing inter-layer exploration in open-boundary models.

\subsection{Comparison with Prior Methods}
\begin{figure*}[t]
    \centering
    \subfloat{\includegraphics[width=0.95\linewidth]{ 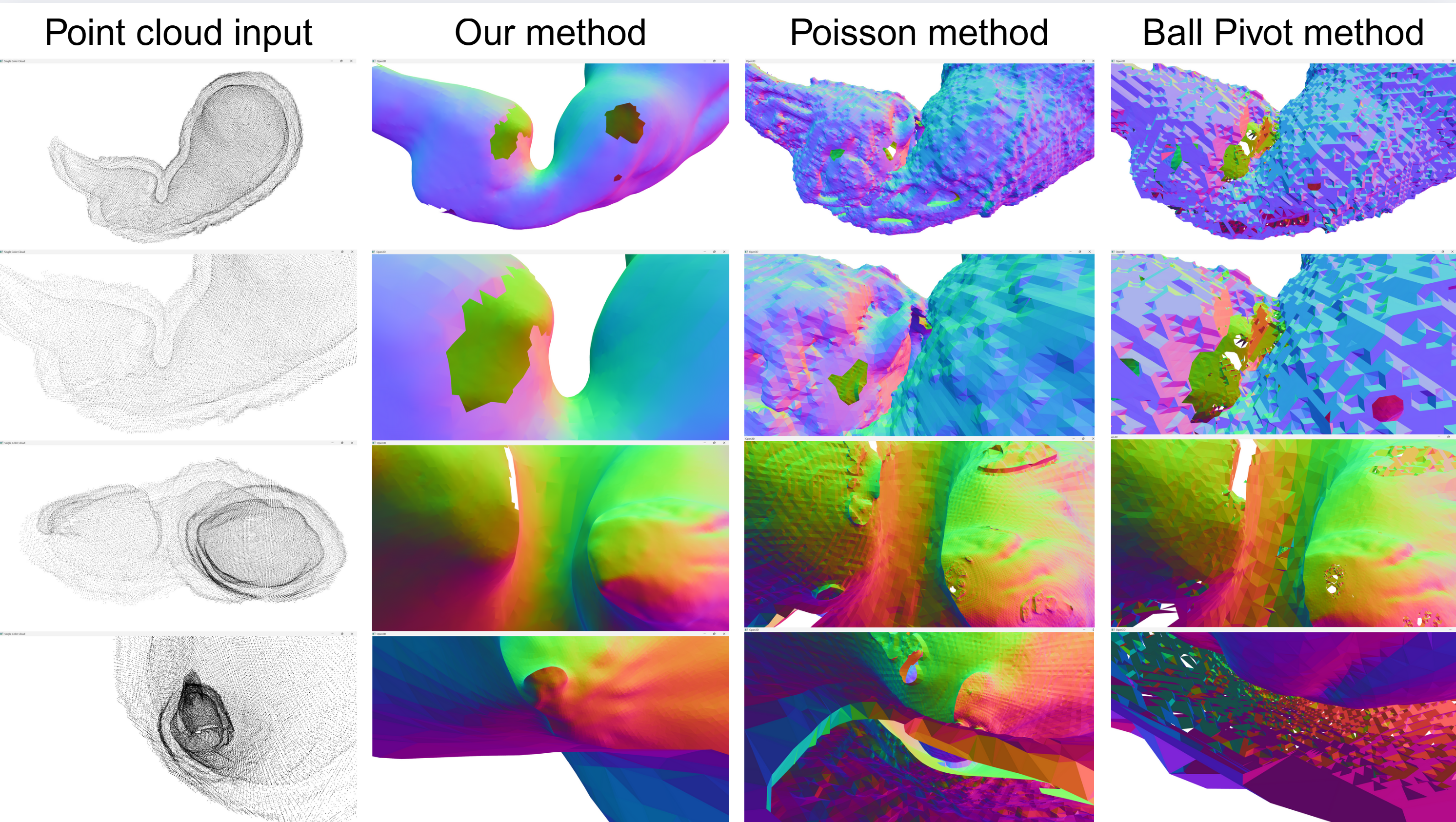 }}
    \caption{Comparison of surface reconstruction methods for the stomach point cloud. \textbf{Our method}: Preserves hole features, generates a clean single-layer mesh, and maintains smooth input geometries. \textbf{Poisson method}: Produces inconsistent normals, double-layered surfaces, erratic merges, imaginary shapes, and irregular textures. \textbf{Ball Pivoting method}: Leads to undesirable hole bridging.}
    \label{fig:MethodContrast}
\end{figure*}

We compare our proposed method for stomach point cloud reconstruction with established approaches, including the Poisson and Ball Pivoting methods, as illustrated in Figure~\ref{fig:MethodContrast}. Our method effectively preserves geometric details and produces a high-quality single-layer mesh, addressing limitations in prior techniques such as inconsistent normals, double layering, and hole bridging.



\subsection{Parallel Acceleration}
To enhance computational efficiency, a parallelized version of the code was developed for the simulation of 3D double layer ball models (composed of 20000 inter points and 20000 outer points) over 500,000 steps. The performance was evaluated on an Intel\textsuperscript{\textregistered} Core\texttrademark{} i9-14900K CPU, featuring 24 physical cores and 32 logical CPUs (with 2 threads per core) running at a maximum clock speed of \SI{6.0}{\giga\hertz}. Tests were conducted across varying numbers of CPU processors, measuring both the time consumed and the processing rate in balls per second. The results, summarized in Table~\ref{tab:parallel_performance}, demonstrate significant improvements in computational speed with increased processor counts, although diminishing returns are observed at higher processor numbers, likely due to overhead in thread management or resource contention.

\begin{table*}[h]
\centering
\setlength{\tabcolsep}{3pt} 
\caption{Performance Metrics for Parallelized Code Execution over 500,000 Steps}
\label{tab:parallel_performance}
\begin{tabular}{c S[table-format=2.0] S[table-format=2.2] S[table-format=4.2] S[table-format=1.5]}
\toprule
\textbf{Implementation} & \textbf{CPU Processors} & \textbf{Time Consumed (s)} & \textbf{Balls per Second} & \textbf{$R_{inter}$} \\
\midrule
Serial  & 1  & 47.06 & 1062.53 & 0.99040 \\
Parallel  & 1  & 47.02 & 1063.48 & 0.99275 \\
Parallel & 2 & 24.90 & 2008.31 & 0.99245 \\
Parallel & 4  & 14.41 & 3470.83 & 0.99215 \\
Parallel & 8 & 9.89  & 5056.12 & 0.99270 \\
Parallel & 12 & 9.11  & 5490.55 & 0.99275 \\
Parallel & 16 & 9.44  & 5294.17 & 0.99215 \\
Parallel & 20 & 10.20 & 4902.58 & 0.99230 \\
\bottomrule
\end{tabular}
\end{table*}

\section{Discussion}
\subsection{Limitation of outer layer false detection}
\label{nonwatertight}
In non-watertight 3D models, the simulation balls may exit the model through surface openings, increasing the likelihood of detecting outer-layer point clouds during the diffusion simulation. Our design incorporates an escape sphere to capture balls that scatter outside the model, thereby reducing unintended collisions with outer-layer points. However, in challenging scenarios, such as a double-layered ball structure with perforations in the inter layer, outer-layer collisions are inevitable, leading to significant false detections that compromise the simulation's effectiveness. And the result of a snail model with holes on the shell is shown in Supplementary Material.


\subsection{Limitation of incomplete inter layer detection}
When there is sharp corners inside the 3D models, the simulation ball can miss this corner, the small radius of the simulation ball can decrease the undetected corners, but it can't totally capture the sharp corners even with large enough steps. 
And the result with sharp corner 3D model will be shown in Supplementary Material.



\section{Conclusion}
We proposed a diffusion-based model that extracts the inter-layer from double-layer point clouds generated by TSDF fusion, enabling its application to complex surfaces. Unlike traditional approaches, which cannot extract the inter-layer from non-watertight 3D models, our method overcomes this limitation through a physics-inspired mechanism. By producing a single-layer model, our approach simplifies the evaluation of completeness and facilitates hole detection for subsequent computer vision processes. Notably, it relies solely on point cloud information, eliminating the need for normal directions or mesh data, resulting in a robust and elegant solution.
\section{Acknowledgement}

\newpage
{\small
\bibliographystyle{ieee_fullname}
\bibliography{egbib}
}

\end{document}